\documentclass[journal]{IEEEtran}

\usepackage[table]{xcolor}
\usepackage{todonotes}
\usepackage{graphicx}
\usepackage{algpseudocode}
\usepackage{algorithm,url}
\usepackage{enumitem,multirow,balance}

\definecolor{Gray}{gray}{0.9}
\definecolor{Gray2}{gray}{0.8}
\definecolor{Gray3}{gray}{0.6}

\newcommand{\bff}[1]{{\textbf{#1}}}

\begin{document}

\title{General Video Game AI: a Multi-Track Framework for Evaluating Agents, Games and Content Generation Algorithms}


\author{Diego Perez-Liebana,~\IEEEmembership{Member,~IEEE,}
         Jialin Liu*,~\IEEEmembership{Member,~IEEE,}
		  Ahmed Khalifa,~
          Raluca D. Gaina,~\IEEEmembership{Student Member,~IEEE,}
          Julian Togelius,~\IEEEmembership{Member,~IEEE,}
          and~Simon~M.~Lucas,~\IEEEmembership{Senior Member,~IEEE} 
 \thanks{D. Perez-Liebana, R. Gaina and S. M. Lucas are with the Department of Electrical Engineering and Computer Engineering (EECS), Queen Mary University of London, London E1 4NS, UK.}
 \thanks{J. Liu is with the Shenzhen Key Laboratory of Computational Intelligence, University Key Laboratory of Evolving Intelligent Systems of Guangdong Province, Department of Computer Science and Engineering, Southern University of Science and Technology, Shenzhen 518055, China. She was with EECS, Queen Mary University of London, London E1 4NS, UK.}
  \thanks{*J. Liu is the corresponding author.} 
  \thanks{A. Khalifa and J. Togelius are with the Department of Computer Science and Engineering, New York University, New York 11201, USA.}
 }

\maketitle
\thispagestyle{empty}
\pagestyle{empty}


\begin{abstract}
General Video Game Playing (GVGP) aims at designing an agent that is capable of playing multiple video games with no human intervention. In 2014, The General Video Game AI (GVGAI) competition framework was created and released with the purpose of providing researchers a common open-source and easy to use platform for testing their AI methods with potentially infinity of games created using Video Game Description Language (VGDL). The framework has been expanded into several tracks during the last few years to meet the demand of different research directions. The agents are required either to play multiple unknown games with or without access to game simulations, or to design new game levels or rules.
This survey paper presents the VGDL, the GVGAI framework, existing tracks, and reviews the wide use of GVGAI framework in research, education and competitions five years after its birth. A future plan of framework improvements is also described.
\end{abstract}

\section{Introduction}

Game-based benchmarks and competitions have been used for testing artificial intelligence capabilities since the inception of the research field. Since the early 2000s a number of competitions and benchmarks based on video games have sprung up. So far, most competitions and game benchmarks challenge the agents to play a single game, which leads to an overspecialization, or overfitting, of agents to individual games. This is reflected in the outcome of individual competitions -- for example, over the more than five years the Simulated Car Racing Competition~\cite{yannakakis2018artificial}\footnote{We cite Yannakakis~\cite{yannakakis2018artificial} and Russell~\cite{russell2016artificial} as standard references for Games and AI (respectively) to reduce the number of non GVGP references.} ran, submitted car controllers got better at completing races fast, but incorporated more and more game-specific engineering and arguably less of general AI and machine learning algorithms. Therefore, this trend threatens to negate the usefulness of game-based AI competitions for spurring and testing the development of stronger and more general AI.




The General Video Game AI (GVGAI) competition~\cite{perez2016general} was founded on the belief that the best way to stop AI researchers from relying on game-specific engineering in their agents is to make it impossible. Researchers would develop their agents without knowing what games they will be playing, and after submitting their agents to the competition all agents are evaluated using an unseen set of games. Every competition event requires the design of a new set of games, as reusing previous games would make this task impossible. 


While the GVGAI competition was initially focused on benchmarking AI algorithms for playing the game, the competition and its associated software has multiple uses. In addition to the competition tracks dedicated to game-playing agents, there are now tracks focused on generating game levels or rules. There is also the potential to use GVGAI for game prototyping, with a rapidly growing body of research using this framework for everything from building mixed-initiative design tools to demonstrating new concepts in game design.

The objective of this paper is to provide an overview of the different efforts from the community on the use of the GVGAI framework (and, by extension, of its competition) for General Game Artificial Intelligence. This overview aims at identifying the main approaches that have been used so far for agent AI and procedural content generation (PCG), in order to compare them and recognize possible lines of future research within this field. The paper starts with a brief overview of the framework and the different competition tracks, for context and completeness, which summarizes work published in other papers by the same authors. The bulk of the paper is centered in the next few sections, which are devoted to discussing the various kinds of AI methods that have been used in the submissions to each track. Special consideration is given to the single-player planning track, as it has existed for longest and received the most submissions up to date. This is followed by a section cataloguing some of the non-competition research uses of the GVGAI software. The final few sections provide a view on the future use and development of the framework and competition: how it can be used in teaching, open research problems (specifically related to the planning tracks), and the future evolution of the competition and framework itself.

\section{The GVGAI Framework}


Ebner et al.~\cite{ebner2013towards} and Levine et al.~\cite{levine2013general} first described the need and interest for such a framework that could accommodate a competition for researchers to tackle the challenge of General Video Game Playing (GVGP). The authors proposed the idea of the Video Game Description Language (VGDL), which later was developed by Schaul~\cite{schaul2013video, schaul2014extensible} in a Python framework for model-based learning and released the first game engine in 2013. Years later, Perez-Liebana et al.~\cite{perez2016general} implemented a version of Schaul's initial framework in Java and organized the first General Video Game AI (GVGAI) competition in 2014~\cite{perez2014gvgpc}, which employed games developed in VGDL. In the following years, this framework was extended to accommodate two-player games~\cite{gainaGVGAI2P,gainaGVGAI2Pb}, level~\cite{khalifa2016general}, rule~\cite{khalifa2017rulegen} generation, and real-world physics games~\cite{perez2017physics}. These competition tracks accumulate hundreds of submissions. Furthermore, the GVGAI Framework and Competition have been used as tools for research and education around the globe, including their usage in taught modules, MSc and PhD dissertation projects (see Section~\ref{sec:edu}).

\begin{figure}[!t]
\centering
\includegraphics[width=.9\linewidth]{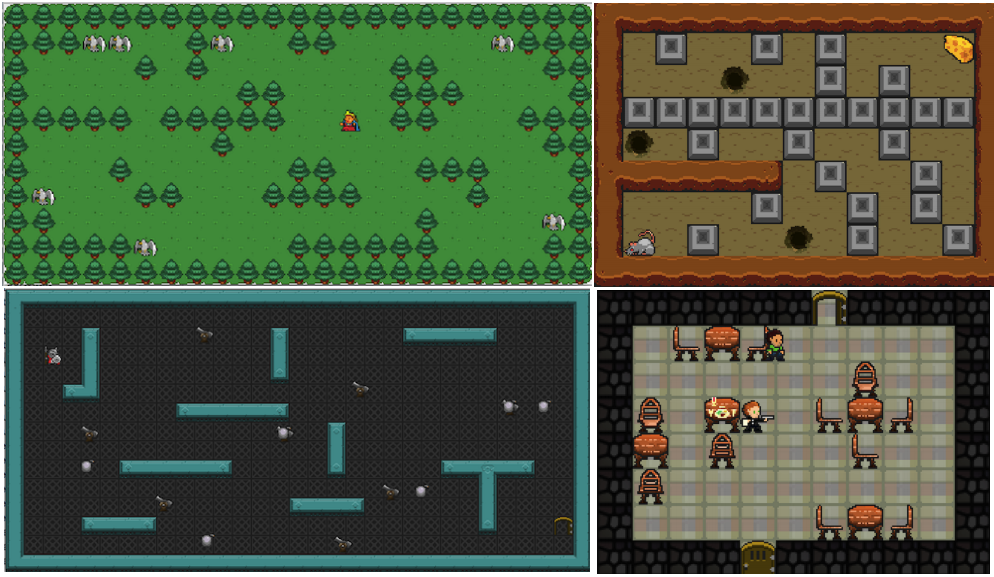}
\caption{\label{fig:vgdlGames}Examples of VGDL games. From top to bottom, left to right: Butterflies, Escape, Crossfire and Wait for Breakfast.}
\end{figure}

VGDL is a text description language that allows for the definition of two-dimensional, arcade, grid-based physics and (generally) stochastic games and levels.
Originally designed for single-player games, the language now admits 2-player challenges. VGDL permits the definition of sprites (objects within the game) and their properties (from speed and behavior to images or animations) in the Sprite Set. Thus this set defines the type of sprites that can take part in the game. Their interactions are regulated in the Interaction Set, which defines the rules that govern the effects of two sprites colliding with each other. 
This includes the specification of score for the games.
The Termination Set defines how the game ends, which could happen due to the presence or absence of certain sprites or due to timers running out. Levels in which the games can be played are defined also in text files. Each character corresponds to one or more sprites defined in the Sprite Set, and the correspondence between sprites and characters is established in the Mapping Set. At the moment of writing, the framework counts on $120$ single-player and $60$ two-player games. Examples of VGDL games are shown in Figure~\ref{fig:vgdlGames}.

VGDL game and level files are parsed by the GVGAI framework, which defines the ontology of sprite types and interactions that are allowed. The benchmark creates the game that can be played either by a human or a bot. For the latter, the framework provides an API that bots (or \textit{agents}, or \textit{controllers}) can implement to interact with the game - hence GVGAI bots can play any VGDL game provided. All controllers must inherit from an abstract class within the framework and implement a constructor and three different methods: \textproc{init}, called at the beginning of every game; \textproc{act}, called at every game tick and must return the next action of the controller; and \textproc{result}, called at the end of the game with the final state. 

The agents do not have access to the rules of the game (i.e. the VGDL description) but can receive information about the game state at each tick. This information is formed by the game status - winner, time step and score -, state of the player (also referred to in this paper as \textit{avatar}) - position, orientation, resources, health points -, history of collisions and positions of the different sprites in the game identified with a unique \textit{type id}. Additionally, sprites are grouped in \textit{categories} attending to their general behavior: Non-Player Characters (NPC), static, movable, portals (which spawn other sprites in the game, or behave as entry or exit point in the levels) and resources (that can be collected by the player). Finally, each game has a different set of actions available (a subset of \textit{left}, \textit{right}, \textit{up}, \textit{down}, \textit{use} and \textit{nil}), which can also be queried by the agent.

In the \textbf{planning} settings of the framework (single-~\cite{perez2014gvgpc} and two-player~\cite{gainaGVGAI2Pb}), the bots can also use a Forward Model. This allows the agent to copy the game state and roll it forward, given an action, to reach a potential next game state. In these settings, controllers have $1$ second for initialization and $40$ms at each game tick as decision time. If the action to execute in the game is returned between $40$ and $50$ milliseconds, the game will play the move \textit{nil} as a penalty. If the agent takes more than $50$ milliseconds to return an action, the bot will be disqualified. This is done in order to keep the real-time aspect of the game. In the two-player case, games are played by two agents in a simultaneous move fashion. Therefore, the forward model requires the agents to also supply an action for the other player, thus facilitating research in general opponent modeling. Two-player games can also be competitive or cooperative, a fact that is not disclosed to the bots at any time.

The \textbf{learning} setting of the competition changes the information that is given to the agents. The main difference with the planning case is that no Forward Model is provided, in order to foster research by learning to play in an episodic manner~\cite{liu2017learningmanual}. This is the only setting in which agents can be written not only in Java, but also in Python, in order to accommodate for popular machine learning libraries written in this language. Game state information (same as in the planning case) is provided in a Json format and the game screen can be observed by the agent at every game tick. Since 2018, Torrado et al.~\cite{Torrado2018} interfaced the GVGAI framework to the OpenAI Gym environment.

The GVGAI framework can also be used for procedural content generation (PCG). In the \textbf{level generation} setting~\cite{khalifa2016general}, the objective is to program a generator that can create playable levels for any game received. In the \textbf{rule generation} case~\cite{khalifa2017rulegen}, the goal is to create rules that allow agents to play in any level received. The framework provides, in both cases, access to the forward model so agents can be used to test and evaluate the content generated.

When generating levels, the framework provides the generator with all the information needed about the game such as game sprites, interaction set, termination conditions and level mapping. Levels are generated in the form of 2d matrix of characters, with each character representing the game sprites at the specific location determined by the matrix. The challenge also allows the generator to replace the level mapping with a new one. When generating rules, the framework provides the game sprites and a certain level. The generated games are represented as two arrays of strings. The first array contains the interaction set, while the second array contains the termination conditions.

As can be seen, the GVGAI framework offers an AI challenge at multiple levels. Each one of the settings (or competition \textit{tracks}) is designed to serve as benchmark for a particular type of problems and approaches. The planning tracks provide a forward model, which favors the use of statistical forward planning and model-based reinforcement learning methods. In particular, this is enhanced in the two-player planning track with the challenge of player modeling and interaction with other another agent in the game. The learning track promotes research in model-free reinforcement learning techniques and similar approaches, such as evolution and neuro-evolution. Finally, the level and rule generation tracks focus on content creation problems and the algorithms that are traditionally used for this: search-based (evolutionary algorithms and forward planning methods), solver (SAT, Answer Set Programming), cellular automata, grammar-based approaches, noise and fractals.

\section{The GVGAI Competition}

For each one of the settings described in the previous section, one or more competitions have been run. All GVGAI competition tracks follow a similar structure: games are grouped in different sets ($10$ games on each set, with $5$ different levels each). \textit{Public} sets of games are included in the framework and allow participants to train their agents on them. For each year, there is one \textit{validation} and one \textit{test} set. Both sets are private and stored in the competition server\footnote{www.gvgai.net; Intel Core i5 machine, 2.90GHz, and 4GB of memory.}. Participants can submit their entries any time before the submission deadline to all training and validation sets, and preliminary rankings are displayed in the competition website (the names of the validation set games are anonymous). 

\subsection{Game Playing Tracks}

In the game playing tracks (planning and learning settings), the competition rankings are computed by first sorting all entries per game according to victory rates, scores and game lengths, in this order.
These per-game rankings award points to the first $10$ entries, from first to tenth position: $25, 18, 15, 12, 10, 8, 6, 4, 2$ and $1$. The winner of the competition is the submission that sums more points across all games in the test set. For a more detailed description of the competition and its rules, the reader is referred to~\cite{perez2014gvgpc}. All controllers are run on the test set after the submission deadline to determine the final rankings of the competition, executing each agent multiple times on each level.


\subsubsection{Planning tracks}

The first GVGAI competition ever held featured the Single-player Planning track in 2014. A full description of this competition can be found at~\cite{perez2014gvgpc}. 2015 featured three legs in a year-long championship, each one of them with different validation and test sets. The Two-player Planning track~\cite{gainaGVGAI2P} was added in 2016, with the aim of testing general AI agents in environments which are more complex and present more direct player interaction~\cite{gainaGVGAI2Pb}. Since then, the single and two-player tracks have run in parallel until 2018.


Table~\ref{tab:allWinnersSPP} shows the winners of all editions up to date, along with the section of this survey in which the method is included and the paper that describes the approach more in depth.

\begin{table}[!t]
\begin{center}
\caption{Winners of all editions of the GVGAI Planning competition. 2P indicates 2-Player track. Hybrid denotes $2$ or more techniques combined in a single algorithm. Hyper-heuristic has a high level decision maker to decides which sub-agent must play (see Section~\ref{sec:spp}). Table extended from~\cite{ashlock2017nofreelunch}.}
\begin{tabular}{|>{\centering\arraybackslash} m{1.7cm}|>{\centering\arraybackslash} m{1.25cm}|>{\centering\arraybackslash} m{2.5cm}|>{\centering\arraybackslash} m{1.25cm}|}
\hline
\textbf{Contest Leg} & \textbf{Winner} & \textbf{Type} & \textbf{Section}\\
\hline
\hline
CIG-14 & OLETS & Tree Search Method & \ref{sec:con-MCTS}~\cite{perez2014gvgpc} \\
\hline
\hline
GECCO-15 & YOLOBOT & Hyper-heuristic & \ref{sec:con-hyper}~\cite{joppen2017informed} \\
\hline
CIG-15 & Return42 & Hyper-heuristic & \ref{sec:con-hyper}~\cite{ashlock2017nofreelunch}\\
\hline
CEEC-15 & YBCriber & Hybrid & \ref{sec:con-hybrids}~\cite{geffner2015width}\\
\hline
\hline
GECCO-16 & YOLOBOT & Hyper-heuristic & \ref{sec:con-hyper}~\cite{joppen2017informed}\\
\hline
CIG-16 & MaastCTS2 & Tree Search Method & \ref{sec:con-MCTS}~\cite{soemers2016cig} \\
\hline
WCCI-16 (2P) & ToVo2 & Hybrid & \ref{sec:2pp:treesearch}~\cite{gainaGVGAI2Pb}\\
\hline
CIG-16 (2P) & Number27 & Hybrid & \ref{sec:2pp:evol}~\cite{gainaGVGAI2Pb}\\
\hline
\hline
GECCO-17 & YOLOBOT & Hyper-heuristic & \ref{sec:con-hyper}~\cite{joppen2017informed} \\
\hline
CEC-17 (2P) & ToVo2 & Hybrid & \ref{sec:2pp:treesearch}~\cite{gainaGVGAI2Pb}\\
\hline
\hline
WCCI-18 (1P)  & YOLOBOT & Hyper-heuristic & \ref{sec:con-hyper}~\cite{joppen2017informed} \\
\hline
FDG-18 (2P)  & OLETS & Tree Search Method & \ref{sec:con-MCTS}~\cite{gainaGVGAI2Pb} \\
\hline
\end{tabular}

\label{tab:allWinnersSPP}
\end{center}
\end{table}

\subsubsection{Learning track}\label{sec:learning}

The GVGAI Single-Player learning track has run for two years: 2017 and 2018, both at the IEEE Conference on Computational Intelligence and Games (CIG). 

In the 2017 edition, the execution of controllers was divided into two phases: learning and validation. In the learning phase, each controller has a limited amount of time, $5$ minutes, for learning the first $3$ levels of each game. The agent could play as many times as desired, choosing among these $3$ levels, as long as the $5$ minutes time limit is respected. In the validation phase, the controller plays $10$ times the levels $4$ and $5$ sequentially. The results obtained in these validation levels are the ones used in the competition to rank the entries. Besides the two sample random agents written in Java and Python and one sample agent using Sarsa written in Java, the first GVGAI single-player learning track received three submissions written in Java and one in Python~\cite{liu2017learning}.
The winner of this track  is a naive implementation of Q-Learning algorithm (Section \ref{sec:qlearning}).

The 2018 edition featured, for the first time, the integration of the framework with the OpenAI Gym API~\cite{Torrado2018}, which results as GVGAI Gym\footnote{\url{https://github.com/rubenrtorrado/GVGAI_GYM}}. This edition also ran with some relaxed constraints. Firstly, only $3$ games are used for the competition, and they are made public. Only $2$ levels for each are provided to the participants for training purposes, while the other $3$ are kept secret and used for computing the final results. Secondly, each agent has an increased decision time of $100ms$. Thirdly, the participants were free to train their agent by themselves using as much time and computational resources as they want before the submission deadline.

This edition of the competition received only 2 entries, \emph{fraBot-RL-QLearning} and \emph{fraBot-RL-Sarsa}, submitted by the same group of contributors from the Frankfurt University of Applied Science.
The results of the entries and sample agents (\emph{random}, \emph{DQN}, \emph{Prioritized Dueling DQN} and \emph{A2C}~\cite{Torrado2018}) are summarized in Table \ref{tab:learningres2018}. For comparison, the planning agent \emph{OLETS} (with access to the forward model) is included. 
\emph{DQN} and \emph{Prioritized Dueling DQN} are outstanding on level 3 (test level) of the game 1, because the level 3 is very similar to the level 2 (training level).
Interestingly, the sample learning agent \emph{DQN} outperformed \emph{OLETS} on the third level of game 1. \emph{DQN}, \emph{Prioritized Dueling DQN} and \emph{A2C} are not applied to the game 3, due to the different game screen dimensions of different levels.  
We would like to refer the readers to \cite{Torrado2018} for more about the GVGAI Gym.


\begin{table}[!t]
\centering
\setlength\tabcolsep{1.5pt}
\caption{\label{tab:learningres2018}
Score and ranking of the submitted agents in the 2018's GVGAI Learning Competition. \textdagger denotes a sample controller.}
\begin{tabular}{c|ccc|ccc|ccc|c}
\hline
\bff{Game} & \multicolumn{3}{c|}{\bff{Game 1}} & \multicolumn{3}{c|}{\bff{Game 2}} & \multicolumn{3}{c|}{\bff{Game 3}} & \multirow{2}{*}{\bff{Ranking}}\\
\bff{Level} & \bff{3} & \bff{4} & \bff{5} & \bff{3} & \bff{4} & \bff{5} & \bff{3} & \bff{4} & \multicolumn{1}{c|}{\bff{5}}\\
\hline
fraBot-RL-Sarsa & -2 & \cellcolor{Gray3}\bff{1} & -1 & 0 & 0 & 0 & \cellcolor{Gray2}\bff{2} & \cellcolor{Gray3}\bff{3} & \cellcolor{Gray2}\bff{2} & 1 \\
fraBot-RL-QLearning & -2 & -1 & -2 & 0 & 0 & 0 & 1 & 0 & \cellcolor{Gray2}\bff{2} & 2\\
Random\textdagger\textdagger & -0.5 & \cellcolor{Gray2}\bff{0.2} & \cellcolor{Gray2}\bff{-0.1} & 0 & 0 & 0 & \cellcolor{Gray3}\bff{3.5} & \cellcolor{Gray2}\bff{0.7} & \cellcolor{Gray3}\bff{2.7} & 3\\
DQN\textdagger & \cellcolor{Gray3}\bff{61.5} & -1 & \cellcolor{Gray3}\bff{0.3} & 0 & 0 & 0 & - & - & - & -\\
Prioritized Dueling DQN\textdagger  & \cellcolor{Gray2}\bff{36.8} & -1 & -2 & 0 & 0 & 0 & - & - & - & -\\
A2C\textdagger & 8.1 & -1 & -2 & 0 & 0 & 0  & - & - & -  & -\\
\hline
OLETS Planning Agent & 41.7 & 48.6 & 3.1 & 0 & 0 & 2.2 & 4.2 & 8.1 & 14 & -\\
\hline
\end{tabular}
\end{table}

\subsection{PCG Tracks}\label{sec:lgt}

In the PCG tracks, participants develop generators for levels or rules that are adequate for any game or level (respectively) given. Due to the inherent subjective nature of content generation, the evaluation of the entries is done by human judges who attend the conference where the competition takes place. For both tracks, during the competition day, judges are encouraged to try pairs of generated content and select which one they liked (one, both, or neither). Finally, the winner was selected based on the generator with more votes.

\subsubsection{Level Generation Track} The first level generation competition was held at the International Joint Conference on Artificial Intelligence (IJCAI) in 2016. This competition received $4$ participants. Each one of them was provided a month to submit a new level generator. Three different level generators were provided in order to help the users get started with the system (see Section~\ref{sec:levelgen} for a description of these). Three out of the four participants were simulation-based level generators while the remaining was based on cellular automata. The winner of the contest was the Easablade generator, a cellular automata described in Section~\ref{sec:lgt:ngram}. The competition was run again on the following year at IEEE CIG 2017. Unfortunately, only one submission was received, hence the the competition was canceled. This submission used a n-gram model to generate new constrained levels using a recorded player keystrokes.

\subsubsection{Rule Generation Track} The Rule Generation track~\cite{khalifa2017rulegen} was introduced and held during CIG 2017. Three different sample generators were provided (Section~\ref{sec:rgt:methods}) and the contest ran over a month's period. Unfortunately, no submissions were received for this track.

\section{Methods for Single Player Planning} \label{sec:spp}

This section describes the different methods that have been implemented for Single Player Planning in GVGAI. All the controllers that face this challenge have in common the possibility of using the forward model to sample future states from the current game state, plus the fact that they have a limited action-decision time. While most attempts abide by the $40$ms decision time imposed by the competition, other efforts in the literature compel their agents to obey a maximum number of calls of the forward model. 

Section~\ref{sec:con-simple} briefly introduces the most basic methods that can be found within the framework. Then Section~\ref{sec:con-MCTS} describes the different tree search methods that have been implemented for this settings by the community, followed by Evolutionary Methods in Section~\ref{sec:con-ea}. Often, more than one method is combined into the algorithm, which gives place to Hybrid methods (Section~\ref{sec:con-hybrids}) or Hyper-heuristic algorithms (Section~\ref{sec:con-hyper}). Further discussion on these methods and their common take-aways has been included in Section~\ref{sec:open}.

\subsection{Basic Methods}\label{sec:con-simple}

The GVGAI framework contains several agents aimed at demonstrating how a controller can be created for the single-player planning track of the competition~\cite{perez2014gvgpc}. Therefore, these methods are not particularly strong.

The simplest of all methods is, without much doubt, \textit{doNothing}. This agent returns the action \textit{nil} at every game tick without exception. The next agent in complexity is \textit{sampleRandom}, which returns a random action at each game tick. Finally, \textit{onesteplookahead} is another sample controller that rolls the model forward for each one of the available actions in order to select the one with the highest action value, determined by a function that tries to maximize score while minimizing distances to NPCs and portals.

\subsection{Tree Search Methods}\label{sec:con-MCTS}

One of the strongest and influential sample controllers is \textit{sampleMCTS}, which implements the Monte Carlo Tree Search (MCTS) algorithm for real-time games. Initially implemented in a \textit{closed loop} version (the states visited are stored in the tree node, without calling forward model during the tree policy phase of MCTS), it achieved the $3^{rd}$ position (out of $18$ participants) in the first edition of the competition.

The winner of that edition, Cou\"etoux, implemented Open Loop Expectimax Tree Search (OLETS), which is an \textit{open loop} (states visited are never stored in the associated tree node) version of MCTS which does not include rollouts and uses Open Loop Expectimax (OLE) for the tree policy. OLE substitutes the empirical average reward by $r_M$,  a weighted sum of the empirical average of rewards and the maximum of its children $r_M$ values~\cite{perez2014gvgpc}.

Schuster, in his MSc thesis~\cite{agent2015torsten}, analyzes several enhancements and variations for MCTS in different sets of the GVGAI framework. These modifications included different tree selection, expansion and play-out policies. Results show that combinations of Move-Average Sampling Technique (MAST) and $n$-Gram Selection Technique (NST) with Progressive History provided an overall higher rate of victories than their counterparts without these enhancements, although this result was not consistent across all games (with some simpler algorithms achieving similar results).

In a different study, Soemers~\cite{soemers2016cig,soemers2016enhancements} explored multiple enhancements for MCTS: Progressive History (PH) and NST for the tree selection and play-out steps, tree re-use (by starting at each game tick with the subtree grown in the previous frame that corresponds to the action taken, rather than a new root node), bread-first tree initialization (direct successors of the root note are explored before MCTS starts), safety pre-pruning (prune those nodes with high number of game loses found), loss avoidance (MCTS ignores game lose states when found for the first time by choosing a better alternative), novelty-based pruning (in which states with features rarely seen are less likely to be pruned), knowledge based evaluation~\cite{perez2014knowledge} and deterministic game detection. The authors experimented with all these enhancements in $60$ games of the framework, showing that most of them improved the performance of MCTS significantly and their all-in-one combination increased the average win rate of the sample agent in $17$ percentage points. The best configuration was the winner of one of the editions of the 2016 competitions (see Table~\ref{tab:allWinnersSPP}).

F. Frydenberg studied yet another set of enhancements for MCTS~\cite{frydenberg2015investigating}. The authors showed that using MixMax backups (weighing average and maximum rewards on each node) improved the performance in only some games, but its combination with reversal penalty (to penalize visiting the same location twice in a play-out) offers better results than vanilla MCTS. Other enhancements, such as macro-actions (by repeating an action several times in a sequence) and partial expansion (a child node is considered expanded only if its children have also been expanded) did not improve the results obtained.

Perez-Liebana et al.~\cite{perez2014knowledge} implemented KB-MCTS, a version of MCTS with two main enhancements. First, distances to different sprites were considered features for a linear combination, where the weights were evolved to bias the MCTS rollouts. Secondly, a Knowledge Base (KB) is kept about how \textit{interesting} for the player the different sprites are, where \textit{interesting} is a measure of curiosity (rollouts are biased towards unknown sprites) and experience (a positive/negative bias for getting closer/farther to beneficial/harmful entities). The results of applying this algorithm to the first set of games of the framework showed that the combination of these two components gave a boost in performance in most games of the first training set.

The work in~\cite{perez2014knowledge} has been extended by other researchers in the field, which also put a special effort on biasing the Monte Carlo (MC) simulations. In~\cite{chu2015biasing}, the authors modified the random action selection in MCTS rollouts by using potential fields, which bias the rollouts by making the agent move in a direction akin to the field. The authors showed that KB-MCTS provides a better performance if this potential field is used instead of the Euclidean distance between sprites implemented in~\cite{perez2014knowledge}. Additionally, in a similar study~\cite{chu2015combining}, the authors substituted the Euclidean distance for a measure calculated by a path-finding algorithm. This addition achieved some improvements over the original KB-MCTS, although the authors noted in their study that using path-finding does not provide a competitive advantage in all games.

Another work by Park and Kim~\cite{park2015mcts} tackles this challenge by i) determining the goodness of the other sprites in the game; ii) computing an Influence Map (IM) based on this; and iii) using the IM to bias the simulations, 
in this occasion by adding a third term to the Upper Confidence Bound (UCB) equation~\cite{yannakakis2018artificial} for the tree policy of MCTS. 
Although not compared with KB-MCTS, the resultant algorithm improves the performance of the sample controllers in several games of the framework, albeit performing worse than these in some of the games used in the study. 

Biasing rollouts is also attempted by dos Santos et al.~\cite{dos2017redundant}, who introduced Redundant Action Avoidance (RAA) and Non-Defeat Policy (NDP); RAA analyzes changes in the state to avoid selecting sequences of actions that do not produce any alteration on position, orientation, properties or new sprites in the avatar. NDP makes the recommendation policy ignore all children of the root node who found at least one game loss in a simulation from that state. If all children are marked with a defeat, normal (higher number of visits) recommendation is followed. Again, both modifications are able to improve the performance of MCTS in some of the games, but not in all.

de Waard et al.~\cite{maarten2016options} introduced the concept of options of macro-actions in GVGAI and designed Option MCTS (O-MCTS). Each option is associated with a goal, a policy and a termination condition. The selection and expansion steps in MCTS are modified so the search tree branches only if an option is finished, allowing for a deeper search in the same amount of time. Their results show that O-MCTS outperforms MCTS in games with small levels or a few number of sprites, but loses in the comparison to MCTS when the games are bigger due to these options becoming too large. 

In a similar line, Perez-Liebana et al.~\cite{perez2017physics} employed macro-actions for GVGAI games that used continuous (rather than grid-based) physics. These games have a larger state space, which in turn delays the effects of the player's actions and modifies the way agents navigate through the level. Macro-actions are defined as a sequence or repetition of the same action during $M$ steps, which is arguably the simplest kind of macro-actions that can be devised. MCTS performed better \textit{without} macro-actions on average across games, but there are particular games where MCTS needs macro-actions to avoid losing at every attempt. The authors also concluded that the length $M$ of the macro-actions impacts different games distinctly, although shorter ones seem to provide better results than larger ones, probably due to a more fine control in the movement of the agents.

Some studies have brought multi-objective optimization to this challenge. For instance, Perez-Liebana et al.~\cite{perez2016multi} implemented a Multi-objective version of MCTS, concretely maximizing score and level exploration simultaneously. In the games tested, the rate of victories grew from $32.24\%$ (normal MCTS) to $42.38\%$ in the multi-objective version, showing great promise for this approach. 
In a different study, Khalifa et al.~\cite{khalifa2017multi} applied multi-objective concepts to evolving parameters for a tree selection confidence bounds equation. A previous work by Bravi~\cite{bravi2016evolving} (also discussed in Section~\ref{sec:con-hybrids}) provided multiple UCB equations for different games. The work in~\cite{khalifa2017multi} evolved, using S-Metric Selection Evolutionary Multi-objective Optimization Algorithm (SMS-EMOA), the linear weights of a UCB equation that results of combining all from~\cite{bravi2016evolving} in a single one. All these components respond to different and conflicting objectives, and their results show that it is possible to find good solutions for the games tested.

A significant exception to MCTS with regards to tree search methods for GVGAI is that of Geffner and Geffner~\cite{geffner2015width} (winner of one of the editions of the 2015 competition, \textit{YBCriber}, as indicated in Table~\ref{tab:allWinnersSPP}), who implemented Iterated Width (IW; concretely IW(1)). IW(1) is a breadth-first search with a crucial alteration: a new state found during search is pruned if it does not make true a new tuple of at most $1$ atom, where \textit{atoms} are Boolean variables that refer to position (and orientations in the case of avatars) changes of certain sprites at specific locations. The authors found that IW(1) performed better than MCTS in many games, with the exception of puzzles, where IW(2) (pruning according to pairs of atoms) showed better performance. This agent was declared winner in the CEEC 2015 edition of the Single-Player Planning Track~\cite{perez2016general}.

Babadi~\cite{babadi2015enhic} implemented several versions of Enforced Hill Climbing (EHC), a breadth-first search method that looks for a successor of the current state with a better heuristic value. EHC obtained similar results to KB-MCTS in the first set of games of the framework, with a few disparities in specific games of the set. 

Nelson~\cite{nelson2016corpus} ran a study on MCTS in order to investigate if, giving a higher time budget to the algorithm (i.e. increasing the number of iterations), MCTS was able to master most of the games. In other words, if the real-time nature of the GVGAI framework and competition is the reason why different approaches fail to achieve a high victory rate. This study provided up to $30$ times more budget to the agent, but the performance of MCTS only increased marginally even at that level. In fact, this improvement was achieved by means of losing less often rather than by winning more games. This paper concludes that the real-time aspect is not the only factor in the challenge, but also the diversity in the games. In other words, increasing the computational budget is not the answer to the problem GVGAI poses, at least for MCTS.

Finally, another study on the uses of MCTS for single player planning is carried out by Bravi et al.~\cite{Bravi2018}. In this work, the focus is set on understanding why and under which circumstances different MCTS agents make different decisions, allowing for a more in-depth description and behavioral logging. This study proposes the analysis of different metrics (recommended action and their probabilities, action values, consumed budget before converging on a decision, etc.) recorded via a \textit{shadow} proxy agent, used to compare algorithms in pairs. The analysis described in the paper shows that traditional win-rate performance can be enhanced with these metrics in order to compare two or more approaches.

\subsection{Evolutionary Methods} \label{sec:con-ea}

The second big group of algorithms used for single-player planning is that of evolutionary algorithms (EA). Concretely, the use of EAs for this real-time problem is mostly implemented in the form of Rolling Horizon EAs (RHEA). This family of algorithms evolves sequences of actions with the use of the forward model. Each sequence is an individual of an EA which fitness is the value of the state found at the end of the sequence. Once the time budget is up, the first action of the sequence with the highest fitness is chosen to be applied in that time step.

The GVGAI competition includes \textit{SampleRHEA} as a sample controller. \textit{SampleRHEA} has a population size of $10$, individual length of $10$ and implements uniform crossover and mutation, where one action in the sequence is changed for another one (position and new action chosen uniformly at random)~\cite{perez2014gvgpc}. 

Gaina et al.~\cite{gaina2017analysis} analyzed the effects of the RHEA parameters on the performance of the algorithm in $20$ games, chosen among the existent ones in order to have a representative set of all games in the framework. The parameters analyzed were population size and individual length, and results showed that higher values for both parameters provided higher victory rates. This study motivated the inclusion of Random Search (\textit{SampleRS}) as a sample in the framework, which is equivalent to RHEA but with an infinite population size (i.e. only one generation is evaluated until budget is consumed) and achieves better results than RHEA in some games.~\cite{gaina2017analysis} also compared RHEA with MCTS, showing better performance for an individual length of $10$ and high population sizes.

Santos et al.~\cite{santos2018improved} implemented three variants for RHEA with shifted buffer (RHEA-SB) by (i) applying the one-step-look-ahead algorithm after the buffer shifting phase; (ii) applying a spatial redundant action avoidance policy~\cite{dos2017redundant}; and (iii) applying both techniques. The experimental tests on $20$ GVGAI single-player games showed that the third variant of RHEA-SB achieved promising results. 

Santos and Bernardino~\cite{santos2018game} applied the avatar-related information, spacial exploration encouraging and knowledge
obtained during game playing to the game state evaluation of RHEA. These game state evaluation enhancements have also been tested on an MCTS agent. The enhancements significantly increased the win rate and game score obtained by RHEA and MCTS on $20$ tested games. 

A different type of information was used by Gaina et al.~\cite{gaina2019sparse-rewards} to dynamically adjust the length of the individuals in RHEA: the flatness of the fitness landscape is used to shorten or lengthen the individuals in order for the algorithm to better deal with sparse reward environments (using longer rollouts for identification of further away rewards), while not harming performance in dense reward games (using shorter rollouts for focus on immediate rewards). However, this had a detrimental effect in RHEA, while boosting MCTS results. Simply increasing the rollout length proved to be more effective than this initial attempt at using the internal agent state to affect the search itself.

A different Evolutionary Computation agent was proposed by Jia et al.~\cite{jia2015gp, jia2015strongly}, which consists of a Genetic Programming (GP) approach. The authors extract features from a screen capture of the game, such as avatar location and the positions and distances to the nearest object of each type. These features are inputs to a GP system that, using arithmetic operands as nodes, determines the action to execute as a result of three trees (horizontal, vertical and action use). The authors report that all the different variations of the inputs provided to the GP algorithm give similar results to those of MCTS, on the three games tested in their study. 

\subsection{Hybrids} \label{sec:con-hybrids}

The previous studies feature techniques in which one technique is predominant in the agent created, albeit they may include enhancements which can place them in the boundary of hybrids. This section describes those approaches that, in the opinion of the authors, would in their own right be considered as techniques that mix more than one approach in the same, single algorithm.

An example of these approaches is presented by Gaina et al.~\cite{gaina2017population}, which analyzed the effects of seeding the initial population of a RHEA using different methods. Part of the decision time budget is dedicated to initialize a population with sequences that are promising, as determined by \textit{onesteplookahead} and \textit{MCTS} agents. Results show that both seeding options provide a boost in victory rate when population size and individual length are small, but the benefits vanish when these parameters are large. 

Other enhancements for RHEA proposed in~\cite{gaina2017enhancements} are incorporating a bandit-based mutation, a statistical tree, a shifted buffer and rollouts at the end of the sequences. The bandit-based mutation breaks the uniformity of the random mutations in order to choose new values according to suggestions given by a uni-variate armed bandit. However, the authors reported that no improvement on performance was noticed. A statistical tree, previously introduced in~\cite{perez2015open}, keeps the visit count and accumulated rewards in the root node, which are subsequently used for recommending the action to take at that time step. This enhancement produced better results with smaller individual length and smaller population sizes. The shifted buffer enhancement provided the best improvement in performance, which consist of shifting the sequences of the individuals of the population one action to the left, removing the action from the previous time step. This variation, similar to keeping the tree between frames in MCTS, combined with the addition of rollouts at the end of the sequences provided an improvement in victory rate ($20$ percentile points over vanilla RHEA) and scores.

A similar (and previous) study was conducted by Horn et al.~\cite{horn2016mcts}. In particular, this study features RHEA with rollouts (as in~\cite{gaina2017enhancements}), RHEA with MCTS for alternative actions (where MCTS can determine any action with the exception of the one recommended by RHEA), RHEA with rollouts and sequence planning (same approach as the shifted buffer in~\cite{gaina2017enhancements}), RHEA with rollouts and occlusion detection (which removes not needed actions in a sequence that reaches a reward) and RHEA with rollouts and NPC attitude check (which rewards sequences in terms of proximity to sprites that provide a positive or negative reward). Results show that RHEA with rollouts improved performance in many games, although all the other variants and additions performed worse than the sample agents. It is interesting to see that in this case the shifted buffer did not provide an improvement in the victory rate, although this may be due to the use of different games.

Schuster~\cite{agent2015torsten} proposed two methods that combine MCTS with evolution. One of them, (1$+$1)-EA as proposed by~\cite{perez2014knowledge}, evolves a vector of weights for a set of game features in order to bias the rollouts towards more interesting parts of the search space. Each rollout becomes an evaluation for an individual (weight vector), using the value of the final state as fitness. The second algorithm is based on strongly-typed GP (STGP) and uses game features to evolve state evaluation functions that are embedded within MCTS. These two approaches join MAST and NST (see Section~\ref{sec:con-MCTS}) in a larger comparison, and the study concludes that different algorithms outperform others in distinct games, without an overall winner in terms of superior victory rate, although superior to vanilla MCTS in most cases.

The idea of evolving weight vectors for game features during the MCTS rollouts introduced in~\cite{perez2014knowledge} (KB-MCTS\footnote{This approach could also be considered an hybrid. Given its influence in other tree approaches, it has also been partially described in Section~\ref{sec:con-MCTS}}) was explored further by van Eeden in his MSc thesis~\cite{van2015analysing}. In particular, the author added A* as a path-finding algorithm to replace the euclidean distance used in KB-MCTS for a more accurate measure and changing the evolutionary approach. While KB-MCTS used a weight for each pair feature-action, being the action chosen at each step by the Softmax equation, this work combines all move actions on a single weight and picks the action using Gibbs sampling. The author concludes that the improvements achieved by these modifications are marginal, and likely due to the inclusion of path-finding. 

Additional improvements on KB-MCTS are proposed by Chu et al.~\cite{chunyin2016position}. The authors replace the Euclidean distance feature to sprites with a grid view of the agent's surroundings, and also the (1$+$1)-EA with a Q-Learning approach to bias the MCTS rollouts, making the algorithm update the weights at each step in the rollout. The proposed modifications improved the victory rate in several sets of games of the framework and also achieved the highest average victory rate among the algorithms it was compared with.

\.{I}lhan and Etaner-Uyar~\cite{ilhan2017mctstdl} implemented a combination of MCTS and \textit{true online} Sarsa ($\lambda$). The authors use MCTS rollouts as episodes of past experience, executing true online Sarsa at each iteration with a $\epsilon$-greedy selection policy. Weights are learnt for features taken as the smallest euclidean distance to sprites of each type. Results showed that the proposed approaches improved the performance on vanilla MCTS in the majority of the $10$ games used in the study. 

Evolution and MCTS have also been combined in different ways. In one of them, Bravi et al.~\cite{bravi2017evolving} used a GP system to evolve different tree policies for MCTS. Concretely, the authors evolve a different policy for each one of the 5 games employed in the study, aiming to exploit the characteristics of each game in particular. The results showed that the tree policy plays a very important role on the performance of the MCTS agent, although in most cases the performance is poor - none of the evolved heuristics performed better than the default UCB in MCTS.

Finally, Sironi et al.~\cite{sironi2018evoApps} designed three Self-Adaptive MCTS (SA-MCTS) that tuned the parameters of MCTS (play-out depth and exploration factor) on-line, using Naive Monte-Carlo, an ($\lambda$, $\mu$)-Evolutionary Algorithm and the N-Tuple Bandit Evolutionary Algorithm (NTBEA)~\cite{kunanusont2017n}. Results show that all tuning algorithms improve the performance of MCTS where vanilla MCTS performs poorly, while keeping a similar rate of victories in those where MCTS performs well. In a follow-up study, however, Sironi and Winands~\cite{Sironi2018b} extend the experimental study to show that online parameter tuning impacts performance in only a few GVGP games, with NTBEA improving performance significantly in only one of them. The authors conclude that online tuning is more suitable for games with longer budget times, as it struggles to improve performance in most GVGAI real-time games.

\subsection{Hyper-heuristics / Algorithm Selection} \label{sec:con-hyper}

Several authors have also proposed agents that use several algorithms, but rather than combining them into a single one, there is a higher level decision process that determines which one of them should be used at each time.

Ross, in his MSc thesis~\cite{ross2014general} proposes an agent that is a combination of two methods. This approach uses A* with Enforced Hill Climbing to navigate through the game at a high level and switches to MCTS when in close proximity to the goal. The work highlights the problems of computing paths in the short time budget allowed, but indicate that goal targeting with path-finding combined with local maneuvering using MCTS does provide good performance in some of the games tested.

Joppen et al.~\cite{joppen2017informed} implemented \textit{YOLOBOT}, arguably the most successful agent for GVGAI up to date, as it has won several editions of the competition. Their approach consists of a combination of two methods: a heuristic Best First Search (BFS) for deterministic environments and MCTS for stochastic games. Initially, the algorithm employs BFS until the game is deemed stochastic, an optimal solution is found or a certain game tick threshold is reached, extending through several consecutive frames if needed for the search. Unless the optimal sequence of actions is found, the agent will execute an enhanced MCTS consistent of informed priors and rollout policies, backtracking, early cutoffs and pruning. The resultant agent has shown consistently a good level of play in multiple game sets of the framework.

Another hyper-heuristic approach, also winner of one of the 2015 editions of the competition (\textit{Return42}, see Table~\ref{tab:allWinnersSPP}), determines first if the game is deterministic or stochastic. In case of the former, A* is used to direct the agent to sprites of interest. Otherwise, random walks are employed to navigate through the level~\cite{ashlock2017nofreelunch}.

Azaria et al.~\cite{Azaria2018} applied GP to evolve hyper-heuristic-based agents. The authors evolved $3$ step-lookahead agents, which were tested on the $3$ game sets from the first 2014 GVGAI competition. The resultant agent was able to outperform the agent ranked at 3rd place in the competition (sample MCTS).

The fact that this type of portfolio agents has shown very promising results has triggered more research into hyper-heuristics and game classification. The work by Bontrager et al.~\cite{bontrager2016matching} used K-means to cluster games and algorithms attending to game features derived from the type of sprites declared in the VGDL files. The resulting classification seemed to follow a difficulty pattern, with $4$ clusters that grouped games that were won by the agents at different rates.

Mendes et al.~\cite{mendes2016hyperheuristic} built a hyper-agent which selected automatically an agent from a portfolio of agents for playing individual game and tested it on the GVGAI framework. This approached employed game-based features to train different classifiers (Support Vector Machines - SVM, Multi-layer Perceptrons, Decision Trees - J48, among others) in order to select which agent should be used for playing each game. Results show that the SVM and J48 hyper-heuristics obtained a higher victory rate than the single agents separately.

Horn et al.~\cite{horn2016mcts} (described before in Section~\ref{sec:con-hybrids}) also includes an analysis on game features and difficulty estimation. The authors suggest that the multiple enhancements that are constantly attempted in many algorithms could potentially be  switched on and off depending on the game that is being played, with the objective of dynamically adapting to the present circumstances. 

Ashlock et al.~\cite{ashlock2017nofreelunch} suggest the possibility of creating a classification of games, based on the performance of multiple agents (and their variations: different enhancements, heuristics, objectives) on them. Furthermore, this classification needs to be stable, in order to accommodate the ever-increasing collection of games within the GVGAI framework, but also flexible enough to allow an hyper-heuristic algorithm to choose the version that better adapts to unseen games.

Finally, Gaina et al.~\cite{Gaina2018} gave a first step towards algorithm selection from a different angle. The authors trained several classifiers on agent log data across $80$ games of the GVGAI framework, in particular obtained only from player experience (i.e. features extracted from the way search was conducted, rather than potentially human-biased game features), to determine if the game will be won or not at the end. Three models are trained, for the early, mid and late game, respectively, and tested in previously not seen games. Results show that these predictors are able to foresee, with high reliability, if the agent is going to lose or win the game. These models would therefore allow to indicate when and if the algorithm used to play the game should be changed. A visualization of these agent features, including win prediction, displayed live while playing games, is available through the Vertig\O~tool~\cite{gaina2018vertigo}, which means to offer better agent analysis for deeper understanding of the agents' decision making process, debugging and game testing.

\section{Methods for Two-Player Planning}

This section approaches agents developed by researchers within the Two-Player Planning setting. Most of these entries have been submitted to the Two-Player Planning track of the competition~\cite{gainaGVGAI2P}. Two methods stood out as the base of most entries received so far, Monte Carlo Tree Search (MCTS) and Evolutionary Algorithms (EA)~\cite{gainaGVGAI2Pb}. On the one hand, MCTS performed better in cooperative games, as well as showing the ability to adapt better to asymmetric games, which involved a role switch between matches in the same environment. EAs, on the other hand, excelled in games with long lookaheads, such as puzzle games, which rely on a specific sequence of moves being identified.

Counterparts of the basic methods described in Section~\ref{sec:con-simple} are available in the framework as well, the only difference being in the One Step Lookahead agent which requires an action to be supplied for the opponent when simulating game states. The opponent model used by the sample agent assumes they will perform a random move (with the exception of those actions that would cause a loss of the game).

\subsection{Tree Search methods}\label{sec:2pp:treesearch}

Most of the competition entries in the first 3 seasons (2016-2018) were based on MCTS (see Section~\ref{sec:con-MCTS}). It is interesting to note that the 2016 winner won again in 2018 -  highlighting the difficulty of the challenge and showing the need for more research focus on multi-player games for better and faster progress.

Some entries employed an Open Loop version of MCTS, which would only store statistics in the nodes of the trees and not game states, therefore needing to simulate through the actions at each iteration for a potentially more accurate evaluation of the possible game states. Due to this being unnecessarily costly in deterministic games, some entries such as \textit{MaasCTS2} and \textit{YOLOBOT} switched to Breadth-First Search in such games after an initial analysis of the game type, a method which has shown ability to finding the optimal solution if the game lasts long enough.

Enhancements brought to MCTS include generating value maps, either regarding physical positions in the level, or higher-level concepts (such as higher values being assigned to states where the agent is closer to objects it hasn't interacted with before; or interesting targets as determined by controller-specific heuristics). The winner of the 2016 WCCI and 2017 CEC legs, \textit{ToVo2}, also employed dynamic Monte Carlo roll-out length adjustments (increased with the number of iterations to encourage further lookahead if budget allows) and weighted roll-outs (the weights per action generated randomly at the beginning of each roll-out).

All agents use online learning in one way or another (the simplest form being the base Monte Carlo Tree Search backups, used to gather statistics about each action through multiple simulations), but only the overall 2016 and 2018 Championship winner, \textit{adrienctx}, uses offline learning on the training set supplied to tune the parameters in the Stochastic Gradient Descent function employed, learning rate and mini batch size. 

\subsection{Evolutionary methods}\label{sec:2pp:evol}

Two of the 2016 competition entries used an EA technique as a base as an alternative to MCTS: Number27 and CatLinux~\cite{gainaGVGAI2Pb}. 

\textit{Number27} was the winner of the CIG 2016 leg, the controller placing 4th overall in the 2016 Championship. Number27 uses a Genetic Algorithm (GA), with one population containing individuals which represent fixed-length action sequences. The main improvement it features on top of the base method is the generation of a value heat-map, used to encourage the agent's exploration towards interesting parts of the level. The heat-map is initialized based on the inverse frequency of each object type (therefore a lower value the higher the object number) and including a range of influence on nearby tiles. The event history is used to evaluate game objects during simulations and to update the value map.

\textit{CatLinux} was not a top controller on either of the individual legs run in 2016, but placed 5th overall in the Championship. This agent uses a Rolling Horizon Evolutionary Algorithm (RHEA). A shift buffer enhancement is used to boost performance, specifically keeping the population evolved during one game tick in the next, instead of discarding it, each action sequence is shifted one action to the left (therefore removing the previous game step) and a new random action is added at the end to complete the individual to its fixed length.

No offline learning was used by any of the EA agents, although there could be scope for improvement through parameter tuning (offline or online).

\subsection{Opponent model}

Most agents submitted to the Two-Player competition use completely random opponent models. Some entries have adopted the method integrated within the sample \textit{One Step Lookahead} controller, choosing a random but non-losing action. In the 2016 competition, \textit{webpigeon} assumed the opponent would always cooperate, therefore play a move beneficial to the agent. \textit{MaasCTS2} used the only advanced model at the time: it remembered Q-values for the opponent actions during simulations and added them to the statistics stored in the MCTS tree nodes; an $\epsilon$-greedy policy was used to select opponent actions based on the Q-values recorded. This provided a boost in performance on the games in the WCCI 2016 leg, but it did not improve the controller's position in the rankings for the following CIG 2016 leg. Most entries in the 2017 and 2018 seasons employed simple random opponent models.

Opponent models were found to be an area to explore further in~\cite{gainaGVGAI2Pb} and Gonzalez and Perez-Liebana looked at $9$ different models integrated within the sample MCTS agent provided with the framework~\cite{gonzalez2017oppmodels}. \textit{Alphabeta} builds a tree incrementally, returning the best possible action in each time tick, while \textit{Minimum} returns the worst possible action. \textit{Average} uses a similar tree structure, but it computes the average reward over all the actions and it returns the action closest to the average. \textit{Fallible} returns the best possible action with a probability $p=0.8$ and the action with the minimum reward otherwise. \textit{Probabilistic} involved offline learning over $20$ games in the GVGAI framework in order to determine the probability of an MCTS agent to select each action, and then using these to determine the opponent action while playing online. \textit{Same Action} returns the same action the agent plays, while \textit{Mirror} returns its opposite. Finally, \textit{LimitedBuffer} records the last $n=20$ actions performed by the player and builds probabilities of selecting the next action based on this data, while \textit{UnlimitedBuffer} records the entire history of actions during the game. When all $9$ opponent models were tested in a round robin tournament against each other, the probabilistic models achieve the highest win rates and two models, \textit{Probabilistic} and \textit{UnlimitedBuffer} outperforming a random opponent model.

Finally, the work done on two-player GVGAI has inspired other research on Mathematical General Game Playing. Ashlock et al.~\cite{Ashlock2018} implemented general agents for three different mathematical coordination games, including the Prisoner's Dilemma. Games were presented at once, but switching between them at certain points, and experiments show that agents can learn to play these games and recognize when the game has changed.

\section{Methods for Single-Player Learning} \label{sec:learning1p}

The GVGAI framework has also been used from an agent learning perspective. In this setting, the agents do not use the forward model to plan ahead actions to execute in the real game. Instead, the algorithms learn the games by repeatedly playing them multiple times (as \textit{episodes} in Reinforcement Learning), ideally improving their performance progressively. 
This section describes first the approaches that tackled the challenge set in the single-player learning track of the 2017 and 2018 competitions, to then move to other approaches.


\subsection{Competition entries}

\subsubsection{Random agent}

A sample random agent, which selects an action uniformly at random at every game tick, is included in the framework (in both Java and Python) for the purposes of testing. This agent is also meant to be taken as a baseline: a learner is expected to perform better than an agent which acts randomly and does not undertake any learning.

\subsubsection{Multi-armed bandit algorithm}

\emph{DontUnderestimateUchiha} by K. Kunanusont is based on two popular Multi-Armed Bandit (MAB) algorithms, $\epsilon$-Decreasing Greedy Algorithm and Upper Confidence Bounds (UCB). At any game tick $T$, the current \emph{best} action with probability $1-\epsilon_T$ is picked, otherwise an action is uniformly randomly selected. The \emph{best} action at time $T$ is determined using UCB with increment of score as reward.
This is a very interesting combination, as the UCB-style selection and the $\epsilon$-Decreasing Greedy Algorithm both aim at balancing the trade-off between exploiting more the best-so-far action and exploring others. Additionally, $\epsilon_0$ is set to $0.5$ and it decreases slowly along time, formalized as $\epsilon_T=\epsilon_0 - 0.0001T$. According to the competition setting, all games will last longer than $2,000$ game ticks, so $\forall T \in \{1,\dots,2000\}$, $0.5 \geq \epsilon_{T}\geq 0.3$. As a result, random decisions are made for approximately $40\%$ time.

\subsubsection{Sarsa}

\emph{sampleLearner}, \emph{ercumentilhan} and \emph{fraBot-RL-Sarsa} are based on the State-Action-Reward-State-Action (Sarsa) algorithm~\cite{russell2016artificial}. The \emph{sampleLearner} and \emph{ercumentilhan} use a subset of the whole game state information to build a new state to reduce the amount of information to be saved and to take into account similar situations. The main difference is that the former uses a square region with fixed size centered at the avatar's position, while the latter uses a first-person view with a fixed distance. \emph{fraBot-RL-Sarsa} uses Sarsa, and it uses the entire screenshot of the game screen as input provided by GVGAI Gym. The agent has been trained using $1000$ episodes for each level of each game, and the total training time was $48$ hours.

\subsubsection{Q-learning\label{sec:qlearning}}

\emph{kkunan}, by K. Kunanusont, is a simple Q-learning~\cite{russell2016artificial} agent using most of the avatar's current information as features, which a few exceptions (such as avatar's health and screen size, as these elements that vary greatly from game to game).
The reward at game tick $t+1$ is defined as the difference between the score at $t+1$ and the one at $t$.
The learning rate $\alpha$ and discounted factor $\gamma$ are manually set to $0.05$ and $0.8$. During the \emph{learning phase}, a random action is performed with probability $\epsilon=0.1$, otherwise, the best action is selected. During the \emph{validation phase}, the best action is always selected. Despite it's simplicity, it won the the first track in 2017.
\emph{fraBot-RL-QLearning} uses the Q-Learning algorithm. It has been trained using $1000$ episodes for each level of each game, and the total training time was $48$ hours.

\subsubsection{Tree search methods}
\emph{YOLOBOT} is an adaption of the \emph{YOLOBOT} planning agent (as described previously in Section~\ref{sec:con-hyper}). As the forward model is no more accessible in the learning track, the MCTS is substituted by a greedy algorithm to pick the action that minimizes the distance to the chosen object at most. According to the authors, the poor performance of \emph{YOLOBOT} in the learning track, contrary to its success in the planning tracks, was due to the collision model created by themselves that did not work well.


\subsection{Other learning agents} \label{subsec:learningagC}

One of the first works that used this framework as a learning environment was carried out by Samothrakis et al.~\cite{samothrakis2015neuroevolution}, who employed Neuro-Evolution in $10$ games of the benchmark. Concretely, the authors experimented with Separable Natural Evolution Strategies (S-NES) using two different policies ($\epsilon$-greedy versus softmax) and a linear function approximator versus a neural network as a state evaluation function. Features like score, game status, avatar and other sprites information were used to evolve learners during $1000$ episodes. Results show that $\epsilon$-greedy with a linear function approximator was the better combination to learn how to maximize scores on each game.

Braylan and Miikkulainen~\cite{braylan2016object} performed a study in which the objective was to learn a forward model on $30$ games. The objective was to learn the next state from the current one plus an action, where the state is defined as a collection of attribute values of the sprites (spawns, directions, movements, etc.), by means of logistic regression. Additionally, the authors transfer the learnt object models from game to game, under the assumption that many mechanics and behaviours are transferable between them. Experiments showed the effective value of object model transfer in the accuracy of learning forward models, resulting in these agents being stronger at exploration.

Also in a learning setting, Kunanusont et al.~\cite{kamolwan2016thesis, kunanusont2017general} developed agents that were able to play several games via screen capture. In particular, the authors employed a Deep Q-Network in $7$ games of the framework of increasing complexity, and included several enhancements to GVGAI to deal with different screen sizes and a non-visualization game mode. Results showed that the approach allowed the agent to learn how to play in both deterministic and stochastic games, achieving a higher winning rate and game score as the number of episodes increased. 

Apeldoorn and Kern-Isberner~\cite{Apeldoorn2017AnAL} proposed a learning agent which rapidly determines and exploits heuristics in an unknown environment by using a hybrid symbolic/sub-symbolic agent model. The proposed agent-based model learned the weighted state-action pairs using a sub-symbolic learning approach. The proposed agent has been tested on a single-player stochastic game, \emph{Camel Race}, from the GVGAI framework, and won more than half of the games in different levels within the first 100 game ticks, while the standard Q-Learning agent never won given the same game length.
Based on \cite{Apeldoorn2017AnAL},
Dockhorn and Apeldoorn~\cite{Dockhorn2018} used exception-tolerant Hierarchical Knowledge Bases (HKBs) to learn the approximated forward model and tested the approach on the 2017 GVGAI Learning track framework, respecting the competition rules. The proposed agent beats the best entry in the learning competition organized at CIG-17~\cite{Dockhorn2018}, but still performed far worse than the best planning agents, which have access to the real forward models.

Using the new GVGAI Gym, Torrado et al.~\cite{Torrado2018} compared 3 implemented Deep Reinforcement Learning algorithms of the OpenAI Gym, Deep Q-Network (DQN), Prioritized Dueling DQN and Advance Actor-Critic (A2C), on 8 GVGAI games with various difficulties and game rules. All the three RL agents perform well on most of the games, however, DQNs and A2C perform badly when no game score is given during a game playing (only win or loss is given when a game terminates). These three agents have been used as sample agents in the learning competition organized at CIG-18.

Finally, Justesen et al.~\cite{justesen2018procedural} implemented A2C within the GVGAI-Gym interface in a training environment that allows learning by procedurally generating new levels. By varying the levels in which the agent plays, the resulting learning is more general and does not overfit to specific levels. The level generator creates levels at each episode, producing them in a slowly increasing level of difficulty in response to the observed agent performance.

\subsection{Discussion} \label{subsec:learningdiscussion}
The presented agents differ between each other in the input game state (Json string or screen capture), the amount of learning time, the algorithm used. Additionally, some of the agents have been tested on a different set of games and sometimes using different game length (i.e., maximal number of game ticks allowed). None of the agents, which were submitted to the 2017 learning competition, using the classic GVGAI framework, have used screen capture.

The Sarsa-based agents performed surprisingly bad in the competition, probably due to the arbitrarily chosen parameters and very short learning time. 
Also, learning 3 levels and testing on 3 more difficult levels given only 5 minutes learning time is a difficult task. An agent should take care of the learning budget distribution and decide when to stop learning a level and to proceed the next one. 

The learning agent using exception-tolerant HKBs~\cite{Dockhorn2018} learns fast. However, when longer learning time is allowed, it is dominated by Deep Reinforcement Learning (DRL) agents. 
Out of the 8 games tested by Torrado et al.~\cite{Torrado2018}, none of the tested 3 DRL algorithms outperformed the planning agents on 6 games. However, on the heavily stochastic game Seaquest, A2C achieved almost double score than the best planning agent, MCTS.

\section{Methods for level generation} \label{sec:levelgen}

Different researchers used different approaches to generate levels for the GVGAI framework. The following subsection describes all known generators either included in the framework or developed during the competition. %




\subsection{Constructive methods}
Constructive generators are designed to generate levels based on general knowledge. For example: enemies should be away from the avatar, walls shouldn't divide the world into islands, etc. Based on the game the generator adjusts a couple of parameters and rules to fit the game as, for example, the number of non-playable characters (NPCs) in the generated level. Constructive generators don't need any simulations after generating the level. The following are the known constructive generators.
\subsubsection{Sample random generator}
This is the most naive method to generate a level. The generator first identifies solid sprites. Solid sprites block the avatar and all NPCs from moving and don't react to anything. The generator adds a selected solid sprite as a border for the generated level to prevent sprites from wandering outside the game screen. Followed by adding one of each character in the level mapping section to a random location in the level. This step ensures the game is playable. Finally, it adds a random amount of random sprites from the level mapping to random locations in the level.
\subsubsection{Sample constructive generator}
This generator uses some general game knowledge to generate the level. First, the generator calculates the level dimensions and the number of sprites in the level, then labels game sprites based on their interactions and sprite types. After that, it constructs a level layout using the solid sprites, to later add the avatar to a random empty location. After knowing the avatar position, the generator adds harmful sprites (those that can kill the avatar) in a far location from the avatar and adds other sprites at any random free locations. Finally, the generator makes sure that the number of goal sprites is sufficient to prevent winning or losing automatically when the game starts.
\subsubsection{Easablade constructive generator}
This is the winner generator for the first level generator competition. The generator is similar to the sample constructive generator but it uses cellular automata to generate the level instead of layering the objects randomly. The cellular automata is run on multiple layers. The first layer is to design the map obstacles, followed by the exit and the avatar, then the goal sprites, harmful sprites, and others.
\subsubsection{N-Gram constructive generator} \label{sec:lgt:ngram}
This generator uses a n-gram model to generate the level. The generator records the player actions from a previous play-through. This action sequence is used to generate the levels using predefined rules and constraints. For example, if the player uses the USE action quite often, the generator will include more enemies in the level. The n-gram is used to specify the rules. Instead of reacting to each separate action, the model reacts to a n-sequence of actions. During the generation process, the algorithm keeps track of the number and position of every generated object to ensure the generated sprites do not overpopulate the level. A single avatar sprite is placed in the lower half of the level.
\subsubsection{Beaupre's constructive pattern generator}~\label{sec:beaupreConstPattern}
In this work, Beaupre et al.~\cite{beaupre2018design} automatically analyzed 97 different games from the GVG-AI framework using a 3x3 sliding window over all the provided GVG-AI levels. They constructed a dictionary of all the different patterns (they discovered $12,941$ unique patterns) with labels about the type of objects in them. The constructive generator starts by checking if the game contain solid sprites (sprites that doesn't allow player to pass through them). If that was the case, the generator fills the edges using border patterns (patterns that contain solid sprites and exists on the edge of the maps). The rest of the game area is filled by random selecting of patterns that maintain the following two heuristics: 1) only one avatar sprite should be found in the level; and 2) all non solid game areas area connected.

\subsection{Search-based methods}
Search-based generators use simulations to make sure the generated level is playable and better than just placing random objects. The following are the known search-based generators.
\subsubsection{Sample genetic generator}
This is a search-based level generator based on the Feasible Infeasible 2 Population Genetic Algorithm (FI2Pop). FI2Pop is a GA which uses 2 populations, one for feasible chromosomes and the other for infeasible chromosomes. The feasible population tries to increase the difference between the OLETS agent (see Section~\ref{sec:con-MCTS}) and one-step look ahead, while the infeasible population tries to decrease the number of chromosomes that violate the problem constraints (i.e. at least one avatar in the game, the avatar must not die in the first 40 steps, etc.). Each population evolves on its own, where the children can transfer between the two populations. This generator initializes the population using sample constructive generator.
\subsubsection{Amy12 genetic generator}
This generator is built on top of the sample genetic generator. The main idea is to generate a level that fits a certain suspense curve. Suspense is calculated at each point in time, by calculating the number of actions that leads to death or tie using the OLETS agent. The algorithm modifies the levels to make sure the suspense curve is not constant during the life time of the game. Good generators are aimed at producing $3$ suspense peeks with values of $50\%$ (where half of the actions, on average lead to losing the game). One of the advantages of using this technique that it makes sure that the generated level is winnable. Games that are hard to win will have a higher peak in the suspense curve, which is not valued highly by the generator.
\subsubsection{Jnicho genetic generator}
This generator~\cite{nichols2016galevelgen} uses a standard GA with similar crossover and mutation operators to the sample GA. The fitness function used is a combination between the score difference and the constraints specified in the sample genetic generator. The score difference is calculated between an Monte Carlo Tree Search agent and One Step Look Ahead agent. The score difference is normalized between $0$ and $1$ to make sure it won't overshadow the constraint values.
\subsubsection{Number13 genetic generator}
This is a modified version of the sample genetic generator. These modifications includes using adaptive crossover mechanism, adaptive mutation rate, a better agent than OLETS, and allowing crossover between feasible and infeasible population, which is not allowed in the sample genetic generator.
\subsubsection{Sharif's pattern generator}
This generator is still work in progress. Sharif et al.~\cite{sharif2017design} identified 23 different patterns by analyzing the grouping of different game sprites from several GVG-AI games. They are working now on using these design patterns as a fitness function for a search based generator.
\subsubsection{Beaupre's evolutionary pattern generator}
Similar to Beaupre's constructive pattern generator in section~\ref{sec:beaupreConstPattern}, they used the constructed dictionary for designing a search based generator. They modified the sample genetic generator provided with the framework to work using patterns instead of using game sprites. They also initialized the generator using the constructive version to speed up the generation process.

\subsection{Constraint-based methods}
\subsubsection{ASP generator}
This generator~\cite{neufeld2015procedural} uses Answer Set Programming (ASP) to generate levels. The main idea is to generate ASP rules that generate suitable levels for the current game. The generated rules consists of three different types. The first type are basic rules, which are based on specific decisions to keep the levels simple (for instance, levels can only have one sprite per tile). The second type are game specific rules, which are extracted from the game description file. An example is the identification of singleton sprites that should only have one sprite in the level. The last type are additional rules to minimize the search space. These rules limit the minimum and maximum number of each sprite type. All the rules are evolved using evolutionary strategy with the algorithm performance difference between \textit{sampleMCTS} and a random agent as the fitness function.

\subsection{Discussion}
The presented generators differ in the amount of time needed to generate a level and the features of the generated content. The constructive generators take the least amount of time to generate a single level without a guarantee that the generated level is beatable. On the other hand, both search-based and constraint-based generators take longer time but generate challenging beatable levels as they use automated playing agents as a fitness function. The constraint-based generator only takes long time to find an ASP generator which could be used to generate many different levels as fast as the constructive generators, while search-based generators take a long time to find a group of similar looking levels.




For the generators that participated in the GVG-AI level generation competition (Easablade, Amy12, Jnicho, and Number13), they have been evaluated during IJCAI 2016 by asking the conference delegates to play two randomly selected levels and choose a preferred one. Each generator was used to generate 3 levels for 4 different games (The Snowman, Freeway, Run, and Butterflies). Easablade was chosen most often ($78.4\%$), followed by Number13, amyP2 and jnicho ($40.3\%$, $39.13\%$ and $34.54\%$, respectively). The winner, Easablade, generated fewer objects than the opponents and nice looking layouts produced by the cellular automata, which is likely is the main reason behind its victory. Most of the generated levels by Easablade, however, were either unbeatable or easy compared to the other generators.


\section{Methods for Rule Generation} \label{sec:rgt:methods}
This section describes the different algorithms that are included in the framework or have been found in the literature~\cite{nielsen2015towards} toward generating rules for the GVG-AI framework.

\subsection{Constructive methods}
Constructive methods are algorithms that generate the rules in one pass without the need to play the game. The constructive methods often incorporate knowledge about game design to generate more interesting games.

\subsubsection{Sample random generator}
This is the simplest generator provided with the framework. The main idea is to generate a game that compiles with no errors. For example, the game shouldn't contain interactions such as killing the end of screen (EOS) sprite. The algorithm starts by generating a random number of interactions by selecting two random sprites (including EOS) and a random interaction rule one by one. The algorithm checks that every interaction is valid before adding it to the generated game. After generating the random interactions, the algorithm generates two termination conditions, one for winning and one for losing. The losing condition is fixed to the avatar being killed, while the winning is either winning the game after a random amount of frames or winning the game when certain sprite count reaches zero.

\subsubsection{Sample constructive generator}
This is a more complex generator that utilizes knowledge about VGDL language and level design to generate more interesting games. The algorithm starts by classifying the game sprites into different categories, such as wall sprites (those that surround the level), collectible/harmful sprites (immovable sprites that cover around $10\%$ of the level), spawner sprites (sprites that spawn another), etc. For each type of sprite, the algorithm has rules to generate interactions based on them. For example, harmful sprites kill the avatar on collision, wall sprites either prevent any movable object from passing through or kill the movable object upon collision, etc. For more details about the rules, the reader is referred to~\cite{khalifa2017rulegen}). After the game interactions are generated, two termination conditions are generated, one for winning and one for losing. The losing condition is fixed to the avatar's death, while the winning condition depends on the current sprites. For example: if collectible sprites exist in the current definition, the winning condition is set to collect them all.

\subsection{Search-based methods}
Search-based methods use a search based algorithm to find a game based on certain criteria that ensure the generated game have better rules than just randomly choosing them. 

\subsubsection{Sample genetic generator}
Similar to the level generation track, the search based algorithm uses FI2Pop to evolve new games. As discussed before, FI2Pop keeps two populations one for feasible games and the other for infeasible games. The infeasible games tries to become feasible by satisfying multiple constraints such as minimizing the number of bad frames (frames contains sprites outside the level boundaries) under certain threshold, the avatar doesn't die in the first 40-frames, etc. On the other hand, the feasible chromosomes try to maximize its fitness. The fitness consists of two parts, the first part is to maximize the difference in performance between the OLETS and MCTS agents, and the difference between MCTS and random agent. The second part is to maximize the number of interaction rules that fires during the simulation of the generated game.
\subsubsection{Thorbjørn generator}
This generator~\cite{nielsen2015towards} is similar to the sample genetic generator. It tries to maximize the difference between the performance of different algorithms. This generator uses evolutionary strategies with mutation and crossover operators to generate an entire game instead of an interaction set and termination conditions.

\subsection{Discussion}
Similar to the level generators, the difference between the different generators is the time used in creation and the features in the output game. The constructive methods take less time but do not guarantee different games or playability, while the search-based generators take long time to generate one game, attempting to satisfy the playability constraints using automated playing agents. \textit{Thorbjorn} is the only generator that creates the whole game, not only the interaction rules and termination conditions, which makes it harder to compare to the rest of the generators. 


The remaining ones are the 3 sample generators that come with the framework, which are compared to each other by doing a user study on the generated games~\cite{khalifa2017rulegen}. The generators are used to generate $3$ new games for $3$ different levels (Aliens, Boulderdash, and Solarfox). The participants in the study were subjected to two generated games by two randomly selected generators and asked to pick the one they prefer. The constructive generator was the preferred one (chosen $76.38\%$ of the time), followed by the genetic ($44.73\%$) and random ($24.07\%$) generators. An explanation for the low preference shown for the genetic generator could be its fitness function: it incorporates a constraint that tries to make sure that the game sprites are always in the playing area. This constraint caused the GA in the current allocated time to favor games that limit considerably the movement of the sprites.



\section{Research that builds on GVGAI} \label{sec:builds}

\subsection{Learn the domain knowledge}
Beside the work relevant to the learning competition, there are some other research work around Reinforcement Learning using the GVGAI framework. 
Narasimhan et al.~\cite{Narasimhan2017} combined a differentiable planning module and a model-free component to a  two-part representation, obtained by mapping the collected annotations for game playings to the transitions and rewards, to speed up the learning. The proposed approach has been tested on 4 GVGAI single-player games and shown its effectiveness on both transfer and multi-task scenarios on the tested games.
The GVGAI Learning Competition  proposes to use a screen-shot of the game screen (at pixel level) at every game tick to represent the current game state. Instead of directly using the screen-shot, Woof and Chen~\cite{Woof2018} used an Object Embedding Network (OEN), which extracted the objects in the game state and compressed object feature vectors (e.g., position, distance to the nearest sprite, etc.) into one single fixed-length feature vector. The DRL agent based on OEN has been evaluated on 5 of the GVGAI single-player games and showed various performance levels on the tested games~\cite{Woof2018}.

\subsection{AI-assisted game design}


Machado et al.~\cite{machado2016shopping} implemented a recommender
system based on the VGDL to recommend game elements, such as sprites and mechanics. Then, the recommender system was expanded to \emph{Cicero}~\cite{machado2017cicero, machado2018cicero}, an AI-assisted game design and debugging tool built on top of the GVGAI. \emph{Cicero} has a statistics tool of the interactions to help figuring out the unused game rules; a visualization system to illustrate the information about game objects and events, a mechanics recommender, a query system~\cite{machado2019kwiri} for in-game data, a playtrace aggregator, a heatmap-based game analysis system and a retrospective analysis application \emph{SeekWhence}~\cite{machado2017seekwhence}.
The gameplay sessions by human players or AI agents can be recorded and every single frame at every game tick can be easily extracted for further study and analysis. 

Recently, Liu et al.~\cite{liu2017evolving} applied a simple Random Mutation Hill Climber (RMHC) and a Multi-Armed Bandit RMHC together with resampling methods to tune game parameters automatically. Games instances with significant skill-depth have been evolved using GVGAI agents.
Furthermore, Kunanusont~et al.~\cite{kunanusont2017n} evolved simultaneously the GVGAI agents as part of the game (opponent models).

Guerrero et al.~\cite{guerrero2017beyondwin} explored five GVGAI agents using four different heuristics separately on playing twenty GVGAI games, allowing different behaviors according to the diverse scenarios presented in the games. In particular, this work explored heuristics that were not focused on winning the game, but to explore the level or interact with the different sprites of the games. These agents can be used to evaluate generated games, thus help evolve them with preferences to particular behaviors.


Khalifa et al.~\cite{khalifa2016modifying} modified MCTS agents by editing the UCT formula used in the agent. Human playing data has been used for modeling to make the modified agents playing in a human-like way. Primary results showed that one of the studied agents achieved a similar distribution of repeated actions to the one by human players. The work was then extended by Bravi et al.~\cite{bravi2017evolving}, in which game-play data have been used to evolve effective UCT alternatives for a specific game. The MCTS agents using new formulas, with none or limited domain information, are compared to a standard implementation of MCTS (the \emph{sampleMCTS} agent of GVGAI) on the game \emph{Missile Command}. Applying the UCT alternatives evolved using game-playing data to a standard MCTS significantly improved its performance.

Besides designing games and the agents used in them, the automatic generation of video game tutorials (aimed at helping players understanding how to play a game) is also an interesting sub-field of study. Green et al.~\cite{green2017spacetofire} pointed out that the GVGAI Framework provides an easy testbed for tutorial generation. The game rules in GVGAI are defined in VGDL, therefore the tutorial generation can be easily achieved by reading and translating VGDL files. 
Further Green et al.~\cite{green2018atdelfi} build a system (AtDelfi) that generates tutorials using the VGDL file and automated AI agents. AtDelfi reads the VGDL file and build a graph of interactions between the game sprites. AtDelfi analyzes the graph to identify the winning path (sequence of nodes starting from player sprite that leads to the winning condition in the graph), losing paths (sequence of nodes starting from the losing condition till there is no dependency), and score path (sequence of nodes starting from player sprite that leads to score change in the graph). These paths are represented as text and videos that explain to the user how to play the game. The text is generated using a string replacement method to generate a human readable instructions, while the videos are recorded using a group of automated agents that won the General Video Game Playing Competition~\cite{perez2014gvgpc} and record every group of frames that cause one of the interactions on the path to trigger.

A more recent work by Anderson et al.~\cite{anderson2018deceptive} focused on designing deceptive games to deceive AI agents and lead the agents away from a globally optimal policy. 
Designing such games helps understand the capabilities and weaknesses of existing AI agents and can serve at a preparation step for designing a meta-agent for GVGP which combines the advantages of different agents.
The authors categorized the deceptions and imported various types of deception to the existing GVGAI games by editing the corresponding VGDL files.
The agents submitted to the GVGAI single-player planing competition have been tested on the new games. Interestingly, the final ranking of the agents on each of the games differed significantly from the rankings in the GVGAI competition. The new designed deceptive games successfully explored the weaknesses of agents which have performed well on the test set of the official competition.

Finally, C. Guerrero-Romero et al., in a vision paper~\cite{Guerrero2018}, proposed a methodology that consists of the use of a team of general AI agents with differentiated skill levels and goals (winning, exploring, eliminating sprites, collecting items, etc.). The methodology is aimed at aiding game design by analyzing the performance of this team of agents as a whole and the provision of logged and visual information that shows the agent experience through the game.



\subsection{Game generation with RAPP}


Nielsen et al.~\cite{nielsen2015general} proposed Relative Algorithm Performance Profile (RAPP) as a measure of relative performance of agents and tested their approach on different general game-playing AI agents using GVGAI framework. The authors showed that well-designed games have clear skill-depth, thus being able to distinct good or bad players. In other words, a strong agent or human player should perform significantly better than a weak agent or human player over multiple playings on well-designed games. For instance, a skillful agent is expected to perform better than a random agent, or one that does not move. 

Then, Nielsen et al.~\cite{nielsen2015towards} integrated the differences of average game scores and win rate between any agent and a random agent to the evaluation of new games either randomly generated or generated by editing existing GVGAI games. 
Though most of the resulted games are interesting to play, there are some exceptions, in which the core challenge of the game has been removed. For instance, the enemy can not heart the player, which makes it no more an enemy. But it still provides useful starting points for human designers.

Kunanusont et al.~\cite{kunanusont2017n} extended the idea of RAPP. 
Five GVGAI agents and a deterministic agent designed for the tested Space Battle Game are used as the candidate opponent, which is considered as part of the game to be evolved.
Two GVGAI agents, One Step Look Ahead (weak), MCTS (strong) and the deterministic agent (mediocre), are used to play multiple times the evolved game for evaluation.
The evaluation function is defined as the minimum of the difference of game scores between the strong and mediocre agents, and the difference of game scores between the mediocre and weak agents, aiming at generating games that can clearly distinguish stronger agents and weak agents.

Recently, Kunanusont et al.~\cite{kunanusont2018scoretrend} used the NTBEA to evolve game parameters in order to model player experience within the game. The authors were able to find parameterizations of three games that, when played by MCTS and RHEA agents, produce predefined and different score trends.


\subsection{Robustness testing}

Perez-Liebana et al.~\cite{perez2016analyzing} ran a study on the winners of the 2014 and 2015 editions of the single player planning competition in order to analyze how robust they were to changes in the environment with regards to actions and rewards. The aim of this work was to analyze a different type of generality: controllers for this framework are developed to play in multiple games under certain conditions, but the authors investigated which could be the effect of breaking those compromises: an inaccurate Forward Model, an agent that does not execute the move decided by the algorithm or score penalties incurred by performing certain actions.

An interesting conclusion on this study is that, once the conditions have been altered, sample agents climb up to the top of the rankings and the good controllers behave worse. Agents that rely on best first search or A* (such as \textit{YOLOBOT} or \textit{Return42}, already described in this paper) handled noise very badly. MCTS also showed to be quite robust in this regard, above other Rolling Horizon agents that could not cope so well with these changes. This work also reinforced the idea that the GVGAI framework and competition are also robust. Despite the changes in the performance of the agents, some controllers do better than others under practically all conditions. The opposite (rankings depending only on noise factors, for instance) would mean that the framework is fragile.

More recently, Stephenson et al.~\cite{stephenson2018continuous} have pointed out that the selection of a proper subset of games for comparing a new algorithm with others is critical, as using a non-suitable representative subset may have a bias to some algorithms. More general, the questions is, given a set of sample problems, how to sample a subset as fair as possible for the algorithms to be tested, and to avoid the bias to any of the algorithms. The authors use an information-theoretic method in conjunction with game playing data to assist in the selection of GVGAI games. Games with higher information gains are used for testing a new agent.

\section{Discussion and open research problems on single- and two-player planning} \label{sec:open}


The single- and two-player planning versions of GVGAI are the ones that have received most attention and research. Despite their popularity and efforts, the best approaches rarely surpass an approximately $50\%$ victory rate in competition game sets, with very low victory rate in a great number of games. Similarly, different MCTS and RHEA variants (including many of the enhancements studied in the literature) struggle to achieve a higher than $25\%$ victory rate in all (more than a hundred) single-player games of the framework. Therefore, increasing performance in a great proportion of games is probably the most challenging problem at the moment.

Literature shows multiple enhancements on algorithms and methods aiming to improve this performance, but in the vast majority of cases the improvements only affect a subset of games or certain configurations of the algorithms. While this is understandable due to the nature of general video game playing, it also shows that the current approaches does not work in order to reach truly general approaches that work across board.

The work described in this survey has shown, however, interesting insights that can point us in the right direction. For instance, several studies show that using more sophisticated (i.e. with A* or other methods such as potential fields) distances to sprites as features works better than Euclidean distances. The downside is that computing these measurements take an important part of the decision time budget, which can't be used in case it is needed for some games or states where the best action to take is not straight-forward. 

In general, one could say that one of the main points to address is how to use the decision time more wisely. Some approaches tried to make every use of the forward model count, like those agents that attempt to learn facts about the game during the rollouts of MCTS. Again, some attempts in this direction have provided marginal improvements, but the problem may be trying to design a general feature \textit{extractor}. In other words, what we try to learn is influenced by what we know about existing games (i.e. some sprites are good, other are bad, some spawn other entities and there are sprites - resources - that the avatar can collect). Some games may require features that have not been thought of, especially because the challenge itself presents games that have not been seen before.

Another improvement that has been tried in several studies is the use of macro-actions (in most cases, a repetition of an action during several consecutive steps) to i) make the action space coarser; and ii) make a better use of the time budget. Again, these modifications have improved performance in certain games (including some that had not been won by any algorithm previously) but they either did not have an impact in others, or they made the performance worse. It is likely that different games can benefit from different macro-action lengths (so work could be done on trying to automatically and dynamically adapt the number of times the action is repeated) but also of more complex structures that allow for high-level planning. In fact, games that require high-level planning are still an open problem to be solved in this setting.

Games classification and the use of hyper-heuristics are also an interesting area for research. Some of the best approaches up to date, as YOLOBOT, do make a differentiation between stochastic and deterministic games to later use one or another algorithm. An open challenge is how to make this classification more accurate and detailed, so an approach could count on a portfolio of (more than 2) algorithms that adapt to every game. Attempts have been made to classify with game features, but results suggest that these classifications and the algorithms used are not strong enough. Devising more general features for this, maybe focused on the agent game-play experience rather than game features, is a line of future research. 

All these unsolved issues apply to both single- and two-player settings, although the latter case adds the difficulty of having an opponent to compete or collaborate with. There are two open problems that arise from this: first, no study has been made that tries to identify the game and behavior of the opponent as collaborative or competitive. Analysis of the other player's intentions can be seen as a sub-field on its own, only that in this case we add the general game playing component to it. Secondly, some advancements have been done in using opponent models that go beyond random, but investigation in more complicated opponent models that better capture and learn the behavior of the other player could potentially yield better results.  

Beside the development of agents for game playing, AI-assisted game design, automatic game testing and game debugging using GVGAI agents have attracted researchers' attention. Some work around evolving game skill-depth using relative performance between GVGAI agents have been done recently, and most of this work has been focused on Relative Algorithm Performance Profiles (RAPP), where \textit{performance} is measured in terms of how well the agents play the given games. However, it is sensible to explore other aspects of agent game-play to influence game design. Factors like the amount of level explored by different agents (so a generator favors those levels or games that allow for a wider exploration, or maybe a progressive one), their decisiveness~\cite{volz2017gameplay} on selecting the best action to take or the entropy of their moves can also be used to this end.

\section{Educational use of GVGAI} \label{sec:edu}

The GVGAI framework has been used to provide engaging assignments for taught modules,
and as the basis for many MSc dissertation projects.  The descriptions below
give an idea of the educational uses of GVGAI but are not intended to
be an exhaustive list.

\subsection{Taught Modules}

GVGAI has been used in at least two distinct ways within taught modules.  
The most typical way is to use design specific aspects of the course around the framework, teaching the students about the main concepts of GVGAI with examples of how to write agents for the selected tracks.  This is then followed up with an assignment, where a significant weight is given to how well each student or group's entry performs in the league.  Several institutions have run private leagues for this, including Otto Von Guericke Universit\"{a}t Magdeburg, University of Essex, University of Muenster, Universidad Carlos III de Madrid, 
Universidad de Malaga and New York University.
Running a private league means the course supervisor has full control over the setup of the league, including when students can enter and how thoroughly the entries are evaluated, and the set of games to evaluate them on.  For the 2-player track, this also allows control over the opponents chosen.
The Southern University of Science and Technology and the Nanjing University have also used GVGAI framework in their AI modules, without running a private league, as assignments when teaching search or reinforcement learning methods.

Another use-case in taught modules is to teach the VGDL part of framework, then set the development of novel and interesting games as the assignment.
This was done to good effect at IT University of Copenhagen, where the students produced a number of challenging puzzle games that were later used in the training and validation sets of the planning track. A similar approach was taken in a module on AI-Assisted Game Design at the University of Essex, where 
planning track games were also produced.

\subsection{MSc Dissertation Projects}


GVGAI offers an extensive range of interesting research challenges, some of which have been addressed in MSc dissertation projects. The majority of the ones we are aware of have focused on the single-player planning track, but this is not surprising as it was the first track to be developed. The single-player planning track
also has the benefit of providing some good sample agents as starting points for further work either in
the sense of extending the sample agents to achieve higher performance, or using the sample agents as a useful source of comparison. A good example is the work on MCTS with options, in which options refer to action sequences designed for specific subgoals. The version with options significantly outperformed the sample MCTS agent on most of the games studied: as with many cases what began as an MSc thesis was later published as a conference paper~\cite{maarten2016options}. 
In our experience this usually provides an excellent educational experience for the student.
Other planning track thesis include \cite{agent2015torsten}, the real-time enhancements of \cite{soemers2016enhancements}, knowledge-based variants~\cite{van2015analysing} and goal-oriented approaches~\cite{ross2014general}.

Beyond the planning tracks, other examples (already described in this survey) include applying Answer-Set Programming (ASP)~\cite{xenija2016thesis} or GAs~\cite{nichols2016galevelgen} to the level generation track and learning from game screen capture~\cite{kamolwan2016thesis}. \cite{kamolwan2016thesis} was essentially a learning track approach before the learning track was running.  
Finally, another approach is to extend the framework in some way, such as developing the two-player learning track~\cite{gaina2016thesis}.

\section{Future Directions}

The GVGAI framework and competition are in constant development. The opportunities that this benchmark provides for different lines of research and education are varied, and this section outlines the future directions planned ahead for the following years.

\subsection{New tracks}

As new challenges are proposed, the possibility of organizing them as competition tracks arise. Below are listed some possible new tracks that can attract interesting research areas.

\subsubsection{Automatic game design}
The game design involves, but not limited to, game generation, level generation, rule generation and play-testing (playing experience, game feeling, fun, etc.), study of game market, user interface design and audio design. The automatic game design becomes an active research topic since the late 2000's. A review of the state of the art in automatic game design can be found in \cite{liu2017evolving}.

A \emph{Game Generation track} would aim at providing AI controllers which automatically generate totally new games or game instances by varying the game parameters, i.e., parameter tuning. How to achieve the former is an open question. The straight-forward way would be providing a particular theme, a database of game objects, or searching spaces of game rules, with which the participants can generate new games. The ideal case would be that the controllers automatically create totally new games from nothing. Though there is a yawning gulf between aspiration and reality, an interdisciplinary field combining automatic game design and domain-specific automatic programming is expected. The latter, automatic game tuning, is relatively easier. Some search-based and population-based methods have been applied to game parameter optimization aiming at maximizing the depth of game variants~\cite{liu2017evolving} or finding more playable games.

\subsubsection{Multi-Player GVGAI}

Multi-agent games has drawn people's attention, for instance, real time strategy games (e.g. StarCraft) and board games (e.g. Mahjong). The study of multi-agent GVGAI is a fruitful research topic. Atari games can also be extended to multi-agent games. In particular, the Pac-Man can be seen as a multi-agent game and related competitions have been held since 2011. The most recent Ms Pac-Man vs Ghost Team Competition~\cite{yannakakis2018artificial}, which included partial observability, was held at CIG in 2016. Nevertheless, a more general multi-agent track is favorable.

The interface of the Two-player Planning Track was initially developed for two or more players, so it has the potential to be expanded to a Multi-player Planning Track, in which an agent is allowed to control more than one player or each of the players is controlled by a separate agent. This future track can be expanded again as a multi-agent learning framework, providing a two-or-more-player learning track.

\subsubsection{Turing Test GVGAI}

Determining if an agent that is playing a game is a human or a bot is a challenge that has been subject of study for many years~\cite{yannakakis2018artificial}, and the idea of applying it to a general video game setting is not new~\cite{lehman2015general}. This offers an interesting opportunity to extend the framework to having a Turing Test Track where participants create AI agents that play like humans for any game that is given. Albeit the understandable difficulty of this problem, the interest for research in this area is significant: what are the features that can make an agent play like a human in any game?


\subsection{General directions}

There are several improvements and additions to the framework that can be done and would potentially affect all existent and future competition tracks. One of these continuous modifications is the constant enlargement of the games library. Not only new games are added for each new edition of the competition, but the work done on automatic game design using the GVGAI framework has the potential to create infinite number of games that can be integrated into the framework.

Adding more games can also be complemented with compatibility with other systems. Other general frameworks like OpenAI Gym~\cite{brockman2016openai}, Arcade Learning Environment (ALE)~\cite{bellemare2013arcade} or Microsoft Malm\"{o}~\cite{johnson2016malmo} count on a great number of single- or multi-player, model-free or model-based tasks. 
Interfacing with these systems would greatly increase the number of available games which all GVGAI agents could play via a common API.
This would also open the framework to 3D games, an important section of the environments the current benchmark does not cover.

With regards to the agents, another possibility is to provide them with a wider range of available actions. For instance, the player could be able to apply more than one action simultaneously, or these actions could form a continuous action space (i.e. pressing a throttle in a range between $0$ and $1$). This would enhance the number of legal combinations for the agent to choose from at each decision step. 

Beside the framework itself, the website for GVGAI could also be improved to provide better and faster feedback to the competition participants. More data analysis features can be added, such as visualization of the score changes during the game playing, the action entropy and the exploration of the game world (heat-map of visited positions). A related work is to provide better and more flexible support for game play metric logging, better support for data mining of results together with visualization, and better data saving, which will help enabling to upload replays (i.e., action logs) from AI agents and human play-throughs.

Another envisaged feature is being able to play the game in a web browser (without any download or installation) by an AI agent or human, and visualize the analyzed features during the game playing in real time. A bonus will be the easy parameterization options for games, thus a player or an AI agent can easily set up the parameters and rules to define the desired game by inserting values directly or generate pseudo-randomly a level to play using some pre-implemented automatic game tuning techniques given some particular goals or features. 

\section{Conclusions}

The GVGAI framework offers the most comprehensive system to date for evaluating the performance of general video game playing agents, and for testing general purpose algorithms for creating new games and creating new content for novel games. The framework has been used in multiple international competitions, and has been used to evaluate the performance of hundreds of general video game agents.

The agent tracks cater for planning agents able to exploit a fast forward model, and learning agents that must learn to react sensibly without the benefits of a forward model.  The planning track already comes in single- and two-player versions; the learning track is currently single-player only, but with a two-player version envisaged.
Although long-term learning may also be used within the planning track, the best-performing agents have, as far as we know, not yet done this. Recent successes in Go indicate what can be achieved by combining learning and planning, so applying a similar system within GVGAI is an interesting prospect. In fact, the combination of different approaches into one is an interesting avenue of future research. And example is the work described in this survey that mixes learning and procedural level generation~\cite{justesen2018procedural}, but one could imagine further synergies such as content generation and learning for two-player games.

The main alternative to GVGAI is the ALE~\cite{bellemare2013arcade}. At the time of writing, ALE offers higher-quality games than GVGAI as they were home-console commercial games of a few decades ago.
In GVGAI terms, ALE offers just two tracks: single-player learning and planning, with the learning track being the more widely used. For future work on machine learning in video games, we predict that the two-player tracks will become the most important, as they offer open-ended challenges based on an arms race of intelligence as new players are developed, and are also outside of the current scope of ALE.
Although ALE has had so far a greater uptake within some sectors of the machine learning community, GVGAI benefits from being much more easily extensible than ALE: it is easy to create new VGDL games, easy to create new levels for these games, and easy to create level generators for them as well. It is also easy to automatically generate variations on existing VGDL games and their levels. This allows for training on arbitrarily large sets of game variations and level variations. In contrast, agents trained on ALE games run a serious risk of overfitting to the game and level they are trained on. An immediate priority is to test the rich set of ALE agents on the equivalent GVGAI-tracks to gain a sense of the relative difficulty of each environment and to learn more of the relative challenges offered by each.

The content creation tracks offer an extremely hard challenge: creating rules or levels for unseen games.  Promising directions include the further development and exploitation of a range of general game evaluation measures~\cite{volz2017gameplay}, and greater use of the best GVGAI agents to perform the play-testing of the  novel rules and levels.

The VGDL has been an important part of GVGAI to date, since it makes it possible to rapidly and concisely specify new games.  However, it is also a source of limitation, as its limited expressiveness makes it hard to make games which are fun for humans to play. VGDL also limits the ease with which complex game mechanics can be embedded in games, which in turn limits the depth of challenge that can be posed for the GVGAI agents. Hence an important future direction is the authoring of GVGAI-compatible games in any suitable language which conform to the necessary GVGAI API in order to ensure compatibility with the desired GVGAI track.

Finally, while the above discussion provides a compelling case for the future of GVGAI as a tool for academic study, we also believe that when it reaches a higher level of maturity it will provide an important tool for game designers.
The vision is to provide an army of intelligent agents with a range of play-testing abilities, and a diverse set of metrics with which to analyze a range of important functional aspects of a game.

\section*{Acknowledgements}

The authors would like to thank the participants of all tracks of the competition for their work and submitted controllers and generators. This work was partially supported by the EPSRC CDT in Intelligent Games and Game Intelligence (IGGI) EP/L015846/1, the Shenzhen Peacock Plan (Grant No. KQTD2016112514355531), the Science and Technology Innovation Committee Foundation of Shenzhen (Grant No. ZDSYS201703031748284) and the Program for University Key Laboratory of Guangdong Province (Grant No. 2017KSYS008).

\bibliographystyle{IEEEtran}
\bibliography{gvgaisurvey}

\end{document}